\documentclass[a4paper,fleqn]{cas-dc}

\usepackage[numbers,sort&compress]{natbib}
\usepackage{capt-of}
\usepackage{array}

\def\tsc#1{\csdef{#1}{\textsc{\lowercase{#1}}\xspace}}
\tsc{WGM}
\tsc{QE}

\begin{document}
\let\WriteBookmarks\relax
\def\floatpagepagefraction{1}
\def\textpagefraction{.001}

\shorttitle{}    

\shortauthors{Anonymous Author}  

\title[mode=title]{CivilBench: Benchmarking Vision-Language Models for Multimodal Information Integration in Civil Engineering}

\tnotemark[funding]

\tnotetext[funding]{This work was supported by the Talent Introduction Program of Sichuan University of Science \& Engineering (Grant No. H40125093).}




\author[1]{Dan Yuan}
\fnmark[1]
\ead{yuan.dan.718@s.kyushu-u.ac.jp}
\credit{Methodology, Software, Formal analysis, Investigation, Data curation, Writing -- original draft, Writing -- review \& editing}

\author[2]{Yue Wen}
\cormark[1]
\fnmark[1]
\ead{wenyue@suse.edu.cn}
\credit{Supervision, Data curation, Writing -- review \& editing}

\author[3]{Zhili He}
\credit{Supervision, Writing -- review \& editing}

\author[2]{Zhenkui Zhou}
\credit{Data curation}

\author[2]{Shicen Liu}
\credit{Data curation}

\author[4]{Ziqiang Zheng}
\cormark[2]
\ead{zhengziqiang1@gmail.com}
\credit{Supervision, Conceptualization, Writing -- review \& editing}


\address[2]{School of Civil Engineering, Sichuan University of Science \& Engineering,
	Zigong 643000, China}

\address[1]{Faculty of Information Science and Electrical Engineering, Kyushu University,
	744 Motooka, Nishi-ku, Fukuoka 819-0395, Japan}

\address[3]{Department of Civil and Environmental Engineering,
	The Hong Kong University of Science and Technology,
	Hong Kong, China}

\address[4]{School of Computer Science and Engineering,
	University of Electronic Science and Technology of China,
	Chengdu 611731, China}

\cortext[1]{Corresponding author.}
\cortext[2]{Corresponding author.}

\fntext[1]{Dan Yuan and Yue Wen contributed equally to this work.}

\begin{abstract}
Artificial intelligence (AI), particularly large language models (LLMs) and vision-language models (VLMs), is increasingly applied in civil and structural engineering for disaster assessment, structural analysis, and engineering decision support. However, the extent to which current VLMs possess the reasoning capabilities required for reliable engineering judgments remains unclear. In this work, we introduce CivilBench, a comprehensive benchmark for evaluating VLMs in civil engineering and disaster-related scenarios. CivilBench covers three major task families: Sequential Reasoning, Image-based Structural Understanding, and Open-world Understanding, encompassing numerical reasoning, structural damage assessment, risk inference, causal analysis, and repair planning. The benchmark is constructed through a systematic pipeline of domain-specific data collection, engineering-oriented captioning, expert annotation, question design, and quality verification. Evaluation of representative state-of-the-art VLMs reveals substantial limitations in domain-specific reasoning, with Sequential Reasoning emerging as a major bottleneck, particularly when models must infer numerical relationships from sparse structured observations. 
Further experiments show that numerical priors and external knowledge can alleviate these limitations, although their effectiveness depends on the model's ability to reliably interpret and use the available evidence.
CivilBench provides a capability-oriented testbed for identifying reasoning limitations and evaluating progress toward reliable AI-assisted engineering decision-making. The benchmark will be publicly available at \url{https://huggingface.co/datasets/traceuryuan/CivilBench}.
\end{abstract}





\maketitle

\section{Introduction}\label{}

Recent advances in large language models (LLMs)~\citep{brown2020language, touvron2023llama} and vision-language models (VLMs) ~\citep{radford2021learning, liu2023visual, zhu2023minigpt} have substantially expanded the capability of artificial intelligence systems to jointly process visual and textual information. By aligning perceptual inputs with high-level semantic reasoning, these models have achieved remarkable performance on a wide range of multimodal tasks, including visual question answering~\citep{antol2015vqa}, image captioning~\citep{you2016image,zheng2023marinegpt}, cross-modal retrieval~\citep{zhen2019deep}, and interactive visual reasoning~\citep{xu2023interactive}. More importantly, the rapid progress of VLMs has opened new opportunities for applying AI to knowledge-intensive and decision-oriented domains~\citep{wong2026marineeval,lu2022learn}, where visual evidence must be interpreted together with contextual and professional knowledge. As a result, VLMs are increasingly being explored in areas such as healthcare ~\citep{singhal2025toward}, marine science~\citep{zheng2024marineinst}, scientific analysis~\citep{bonamente2017statistics}, industrial inspection ~\citep{zhang2024vision}, and decision support systems~\citep{sprague1980framework}.

Civil and structural engineering represents one of the most important yet underexplored domains for VLM-based intelligence. Engineering practice often involves interpreting complex visual observations~\citep{xue2026self}, analyzing structural conditions~\citep{park2025anti}, assessing potential risks~\citep{hei2026large}, and making decisions under uncertainty~\citep{hei2025capturing}. 
Reliable VLMs have the potential to support infrastructure inspection~\citep{lam2026large},
post-disaster assessment~\citep{chen2026integration},
and structural monitoring~\citep{ren2026pretrained,xu2025image}.
By integrating visual, textual, and quantitative information, these models may improve the efficiency and consistency of engineering assessment and decision-making~\citep{erfani2026applications}.
Such capabilities are especially valuable in safety-critical scenarios, where timely and accurate analysis can support disaster response, reduce inspection costs, and enhance the resilience of infrastructure systems. However, whether current VLMs can truly perform as reliable civil engineering assistants remains an open question~\citep{sharma2026applications, benfenati2025foundation}. Unlike general multimodal tasks, civil engineering problems require professional domain knowledge, physical intuition, numerical reasoning, and mechanism-level understanding of materials, loads, damage evolution, and structural behavior.

\begin{figure*}[t]
	\centering
	\includegraphics[width=\textwidth]{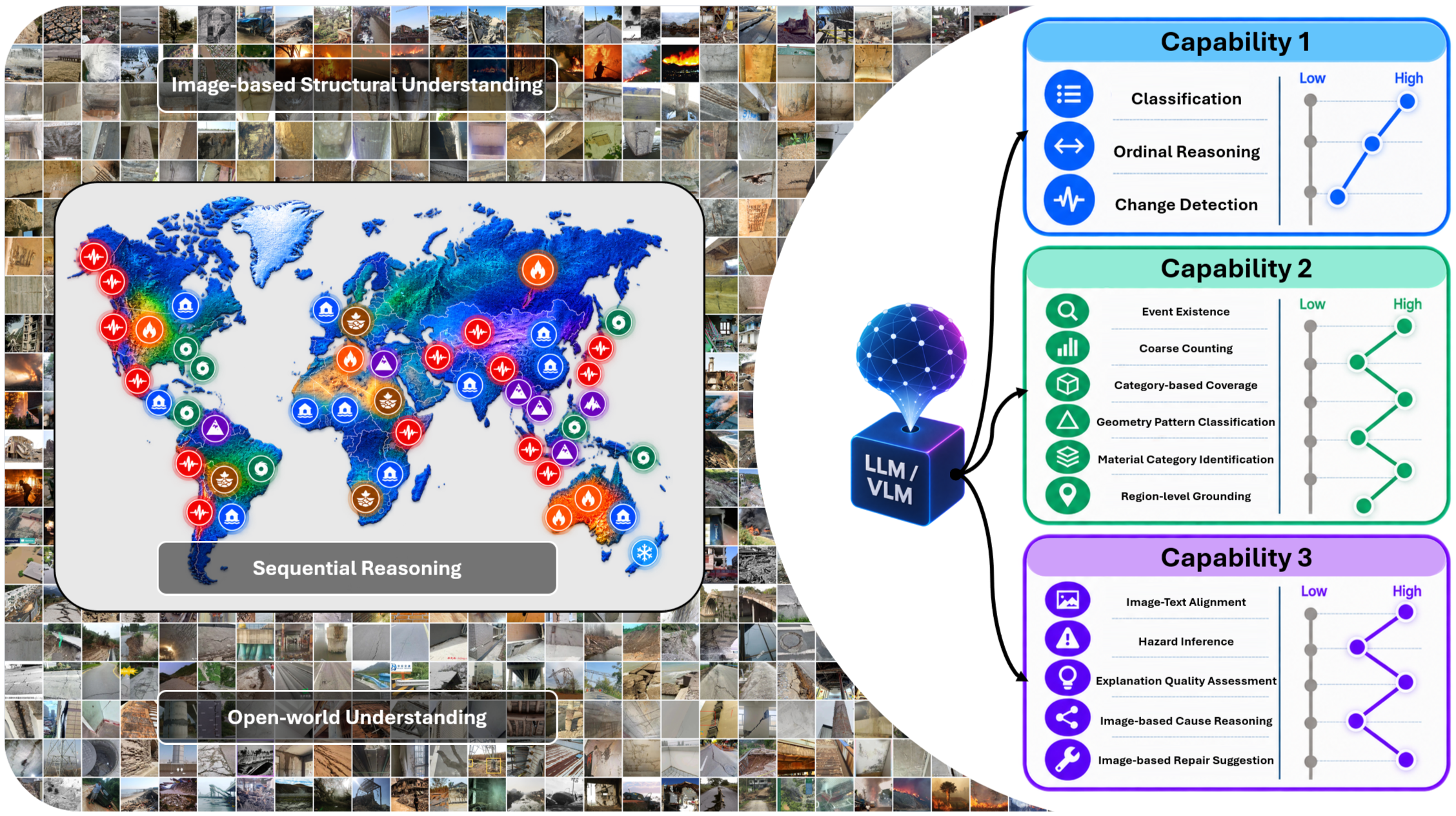}
	
	\caption{Overview of CivilBench and its capability-oriented evaluation framework for vision-language models in civil engineering.}
	
	\label{fig:CivilBench_overview}
\end{figure*}

Despite the growing interest in applying multimodal AI to specialized domains, existing VLM evaluations remain insufficient for civil and structural engineering. Most established benchmarks focus on general-purpose multimodal understanding, including visual question answering~\citep{antol2015vqa}, compositional visual reasoning~\citep{hudson2019gqa}, and multidisciplinary multimodal understanding~\citep{yue2023mmmu}.
These evaluations rarely capture the requirements of engineering practice, such as \textit{interpreting damage severity}, \textit{identifying failure mechanisms}, \textit{reasoning over temporal structural responses}, \textit{estimating quantitative indicators}, and \textit{recommending appropriate repair or mitigation actions}. Moreover, engineering scenarios frequently involve noisy and heterogeneous visual data, partial occlusions, complex spatial layouts, domain-specific terminology, and physical constraints that are absent from generic datasets. Consequently, strong performance on existing benchmarks does not necessarily imply professional reliability in civil engineering contexts~\citep{yeum2015vision}. This mismatch leaves a critical gap: we still lack a rigorous and domain-specific benchmark for understanding whether VLMs can support trustworthy engineering reasoning and decision-making.

Directly evaluating VLMs in civil engineering is challenging because meaningful assessment must go beyond visual recognition~\citep{kampelopoulos2025review}. A model may correctly identify cracks, corrosion, or deformation from the visual observations, but still fail to infer the underlying cause, evaluate the associated risk, or suggest a feasible repair strategy. Similarly, numerical answers in engineering are not isolated values; they often reflect physical consistency, temporal evolution, structural constraints, and professional judgment. Therefore, a useful and practical benchmark should not only test whether a model can describe 1) what is visible, but also 2) whether it can reason about why it occurs, 3) how severe it is, 4) what consequences it may have, and 5) what actions should be taken. Such evaluation is essential for distinguishing superficial multimodal understanding from genuine engineering intelligence.

To address this need, we introduce CivilBench, the first large-scale benchmark specifically designed to evaluate VLMs in civil and structural engineering scenarios. As illustrated in Figure~\ref{fig:CivilBench_overview}, CivilBench provides a comprehensive testbed covering diverse visual, structured, and knowledge-intensive evaluation scenarios. It spans multiple task families and capability dimensions, including sequential numerical reasoning, temporal analysis, structural damage understanding, risk inference, failure-cause analysis, mechanism-level interpretation, and repair or mitigation suggestion. By integrating both fundamental perception and open-world expert reasoning, CivilBench enables a systematic assessment of the gap between general multimodal understanding and professional engineering judgment. Beyond serving as a benchmark for model comparison, CivilBench is intended to support the broader development of AI-assisted civil engineering systems, including infrastructure inspection, disaster assessment, structural health monitoring, safety evaluation, and intelligent decision support.

A key strength of CivilBench lies in its scientifically grounded and comprehensive dataset construction process. Rather than simply collecting image-question pairs, we design a structured pipeline that emphasizes engineering validity, annotation reliability, and practical relevance. The construction process includes domain-specific data collection, engineering-oriented captioning, task-driven question generation, expert-guided annotation, and multi-stage quality verification. Each sample is designed to reflect realistic engineering scenarios and professional reasoning workflows, ensuring that the benchmark evaluates not only visual understanding but also domain-specific inference and decision-making. This careful design allows CivilBench to cover a wide range of engineering conditions, task difficulties, and reasoning types, making it both practically meaningful and analytically informative.

We further develop unified and scalable evaluation protocols for heterogeneous task types, including quantitative prediction, ordinal comparison, damage assessment, risk inference, and mechanism-level reasoning. These protocols are designed to be both scientifically solid and intuitively interpretable, enabling fair comparison across different VLMs while providing detailed diagnostic insights into their strengths and weaknesses. Through extensive benchmarking of representative state-of-the-art VLMs, we find that current models still struggle with domain-specific civil engineering tasks. In particular, they exhibit limitations in physical reasoning, numerical prediction, temporal understanding, failure-mode interpretation, and reliable decision support. These findings highlight the substantial gap between general-purpose VLMs and trustworthy engineering expertise, and point to important directions for future research in domain-aware multimodal intelligence. Our contributions are summarized as follows:

\begin{itemize}
	\item \textbf{Capability-oriented evaluation.}
	We introduce CivilBench, a benchmark that moves beyond generic image--text understanding to systematically assess the engineering reasoning capabilities of VLMs. Rather than relying solely on aggregate accuracy, CivilBench enables fine-grained diagnosis of how models interpret structured observations, understand structural conditions, and make domain-grounded judgments.
	
	\item \textbf{Unified engineering benchmark.}
	CivilBench integrates three task families and 14 capability dimensions, spanning sparse numerical and sequential reasoning, image-based structural understanding, and open-world engineering inference. Built through domain-specific data collection, expert annotation, and quality verification, it provides a unified testbed for examining the gap between multimodal perception and engineering reasoning.
	
	\item \textbf{Reasoning bottlenecks and evidence utilization.}
	Extensive evaluation of representative VLMs identifies Sequential Reasoning as a major bottleneck, particularly in inferring numerical relationships from limited observations. Further experiments show that numerical priors and external knowledge can improve performance, while exposing the challenges of reliably aligning retrieved evidence with engineering tasks.
\end{itemize}

The remainder of this paper is organized as follows.
Section~\ref{RW} reviews related work on VLMs, multimodal benchmarks, and civil engineering datasets.
Section~\ref{CB} introduces the design, construction, and evaluation framework of CivilBench.
Section~\ref{Exp} presents the experimental setup, benchmarking results, and further analysis.
Section~\ref{DL} discusses the main findings and limitations.
Finally, Section~\ref{Con} summarizes the study and outlines future research directions.

\section{Related Work}
\label{RW}

\subsection{VLMs for Understanding and Reasoning}
Recent VLMs have advanced multimodal reasoning by coupling visual encoders with LLMs, enabling open-ended responses grounded in images. They perform well on tasks such as visual question answering~\citep{antol2015vqa}, captioning~\citep{anderson2018bottom}, retrieval~\citep{radford2021learning}, and interactive reasoning~\citep{liu2023visual}, showing strong ability to align visual perception with semantic knowledge. However, these capabilities are mostly demonstrated on generic visual concepts, commonsense reasoning, and web-scale image-text associations~\citep{radford2021learning}. Civil and structural engineering present a more demanding setting~\citep{yeum2015vision, cha2018autonomous}. Engineering tasks require recognizing subtle structural conditions, comparing damage evolution, interpreting quantitative information, and linking observations to physical mechanisms such as material degradation, load transfer, and failure modes~\citep{koch2015review, kampelopoulos2025review}. These requirements go beyond visual recognition and demand domain knowledge, numerical consistency, and safety-aware reasoning. Since most VLMs are trained on broad web data with limited engineering supervision, they may generate plausible but physically inconsistent or professionally unreliable responses. This motivates dedicated evaluation of whether VLMs can support mechanism-aware and decision-oriented reasoning in civil engineering.

\subsection{Benchmarking VLMs}
Evaluating VLMs is challenging because their outputs are often open-ended, multimodal, and sensitive to prompts. Benchmarks such as  VQAv2~\citep{antol2015vqa}, GQA~\citep{hudson2019gqa}, and ScienceQA~\citep{lu2022learn} support structured evaluation, while SEED-Bench~\citep{li2024seed}, MMMU~\citep{yue2023mmmu}, and MMBench~\citep{liu2024mmbench}) assess broader multimodal capabilities.
However, these benchmarks are mainly designed for general-purpose evaluation and do not fully reflect civil engineering requirements. Civil engineering evaluation requires more than semantic correctness. Model responses must also be numerically consistent, physically plausible, temporally coherent, and useful for decision-making. A model may identify visible damage but fail to infer its severity, cause, risk, or proper intervention. Moreover, engineering tasks involve heterogeneous formats, including numerical prediction, ordinal comparison, visual diagnosis, and open-ended expert reasoning, making unified evaluation difficult~\citep{zhang2025mme}. CivilBench addresses these challenges through capability-oriented task design and standardized evaluation protocols, enabling systematic and intuitive assessment of VLMs in sequential reasoning, structural understanding, and open-world engineering decision-making.

\subsection{Civil and Structural Datasets}

Existing civil and structural engineering datasets mainly focus on perception tasks such as crack detection~\citep{zou2018deepcrack, kulkarni2022crackseg9k}, corrosion recognition~\citep{nabizadeh2023automated},
pavement distress classification~\citep{majidifard2020pavement}, bridge inspection~\citep{dorafshan2018sdnet2018, bianchi2021coco},
and post-disaster damage assessment~\citep{gupta2019creating, rahnemoonfar2023rescuenet}. These datasets typically provide image-level labels, bounding boxes, segmentation masks, or damage categories, and have been useful for training computer vision models to identify visible defects~\citep{maeda2018road}. Recent multimodal benchmarks introduce textual descriptions or
question-answer pairs, but their evaluations mainly address descriptive understanding~\citep{lu2022learn}, object recognition~\citep{liu2024mmbench},
or attribute queries~\citep{goyal2017making}.

Despite recent advances, two key research gaps remain.
First, existing benchmarks lack a systematic assessment of VLM capabilities in civil and structural engineering,
particularly in quantitative reasoning, engineering inference, and multimodal alignment. Current civil engineering datasets
primarily focus on damage recognition, with limited coverage of higher-level engineering tasks.
Second, existing evaluations are often limited to individual tasks, making it difficult to compare model performance
across different engineering capabilities and identify their specific strengths and weaknesses.
To address these gaps, CivilBench integrates visual and quantitative tasks within a unified evaluation framework, supported by
visual necessity testing, expert verification, and standardized evaluation protocols. By identifying capability limitations
across engineering tasks, CivilBench also provides insights for future VLM development in civil engineering.

\section{CivilBench}
\label{CB}

This section introduces the design and construction of CivilBench. We first describe the benchmark design principles and the dataset construction process, including data collection, annotation, and quality control. We then present the task taxonomy and evaluation dimensions used to assess different engineering capabilities. Finally, we report the dataset statistics, task distribution, and image characteristics.

\subsection{Benchmark Design Principles}\label{sec31}

CivilBench is designed to assess VLMs under realistic civil and structural engineering constraints, where reliable reasoning requires visual evidence, domain consistency, and objective evaluation. Our construction is motivated by the limitations of generic multimodal benchmarks, which may overestimate model capability through language priors, ambiguous questions, or subjective scoring. To provide a scientifically grounded benchmark, CivilBench follows three key principles.
\begin{itemize}
    \item \textbf{Visual grounding}. Each question must rely on visual inputs, such as structural conditions, damage patterns, component states, or surrounding context. Text-only validation identifies and removes samples answerable without visual evidence, ensuring that CivilBench evaluates genuine image-based engineering reasoning rather than linguistic shortcuts.
    \item \textbf{Evaluation clarity}. To reduce ambiguity, evaluation tasks are formulated with deterministic ground truth and constrained outputs, including classification, ranking, ordinal comparison, and multiple-choice formats. This design preserves engineering relevance while enabling objective and consistent comparison across models.
    \item \textbf{Reproducibility}. Although expert knowledge is incorporated during annotation and verification, model evaluation relies only on fixed labels and predefined metrics. This removes variability from human scoring and supports stable benchmarking across models and experimental settings.
\end{itemize}
These three designed principles ensure that CivilBench is visually grounded, professionally relevant, and reproducible, while establishing the methodological basis for the dataset construction pipeline and task-specific evaluation protocols presented in the following sections.

\subsection{Dataset Construction}
\label{sec32}
\begin{figure*}[t]
    \centering
    \includegraphics[width=0.85\textwidth]{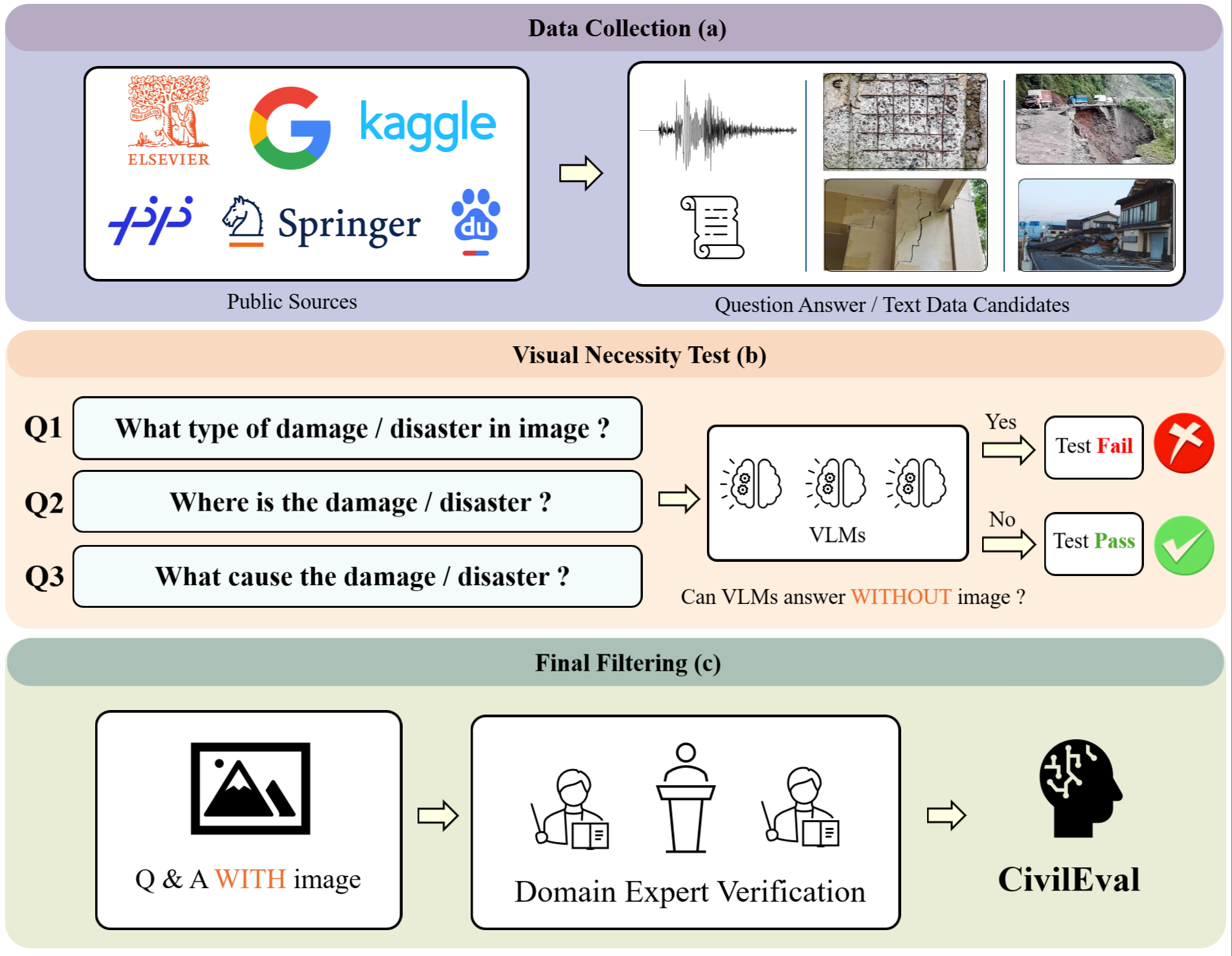}
    
    \caption{Overview of the CivilBench construction pipeline: (a) data collection, (b) visual necessity test, and (c) final filtering.}
    
    \label{fig1}
\end{figure*}

Following the above principles, CivilBench is constructed through a multi-stage pipeline that prioritizes engineering validity, visual dependence, and evaluation consistency. The overall construction workflow is illustrated in Figure~\ref{fig1}. Rather than merely collecting image–question pairs, the dataset evaluates whether models can reason from structural evidence in realistic civil engineering scenarios.

\textbf{Data collection.} Figure~\ref{fig1}(a) illustrates the initial data collection stage. CivilBench integrates heterogeneous data from both public and private sources, including raw images and structured engineering records, to support both sequential reasoning and image-based structural understanding. For Sequential Reasoning Tasks, public tabular and textual resources describing hazard, exposure, and risk-related variables are combined with project-specific experimental records. These data are organized into pseudo-temporal sequences to support classification, ordinal comparison, and change detection. All data were manually screened and annotated by the authors. Publicly available data are used in accordance with their applicable licenses and terms of use. For third-party data without explicit redistribution permission, only derived annotations and metadata will be released where permitted.

For image-based tasks, we collect samples from structural inspection datasets, disaster-related datasets, and publicly available web images using curated civil engineering keywords. The images cover diverse scenarios, including cracking, spalling, collapse, slope failure, and other infrastructure damage, with variations in scale, viewpoint, scene complexity, and image quality. This design introduces realistic visual heterogeneity and domain shift beyond conventional natural-image benchmarks.

\textbf{Question construction and annotation}. CivilBench follows two aligned annotation protocols. For Sequential Reasoning Tasks, records are grouped into short context-query sequences. The context contains labeled observations, while the query requires predicting a target label, comparing severity, or detecting change. This ``context-to-query'' formulation encourages relational and temporal reasoning rather than reliance on isolated label priors. For multiple-choice visual question answering (MC-VQA) tasks, each image is paired with domain-specific questions and four mutually exclusive answers. Questions cover predefined capabilities, including event recognition, damage assessment, geometry analysis, material identification, region grounding, hazard inference, cause analysis, and repair recommendations. Consistent engineering terminology ensures that answers are visually grounded and objectively verifiable.

\textbf{Visual necessity test.} The visual necessity test is illustrated in Figure~\ref{fig1}(b). To prevent language-only shortcuts, all image-based question-answer candidates are evaluated by removing the associated image while retaining the question and answer options. Samples that can be answered correctly without visual information are discarded. This filtering step ensures that the remaining samples genuinely require visual evidence, reducing the influence of prior knowledge, textual cues, or option-level bias.

\textbf{Expert verification and final formatting.}
The final stage of the pipeline is illustrated in Figure~\ref{fig1}(c). After visual necessity filtering, structural engineering experts review each sample for correctness, clarity, and engineering consistency. Ambiguous, misleading, or technically inaccurate samples are revised or removed. The remaining samples are then organized into a unified format consisting of the input image or sequence, question, candidate answers, and ground-truth label. This process ensures a consistent benchmark for reliable evaluation of VLMs in civil engineering.

\subsection{Capability-Oriented Task Taxonomy}

\begin{figure*}[!t]
	\centering
	\includegraphics[width=\textwidth]{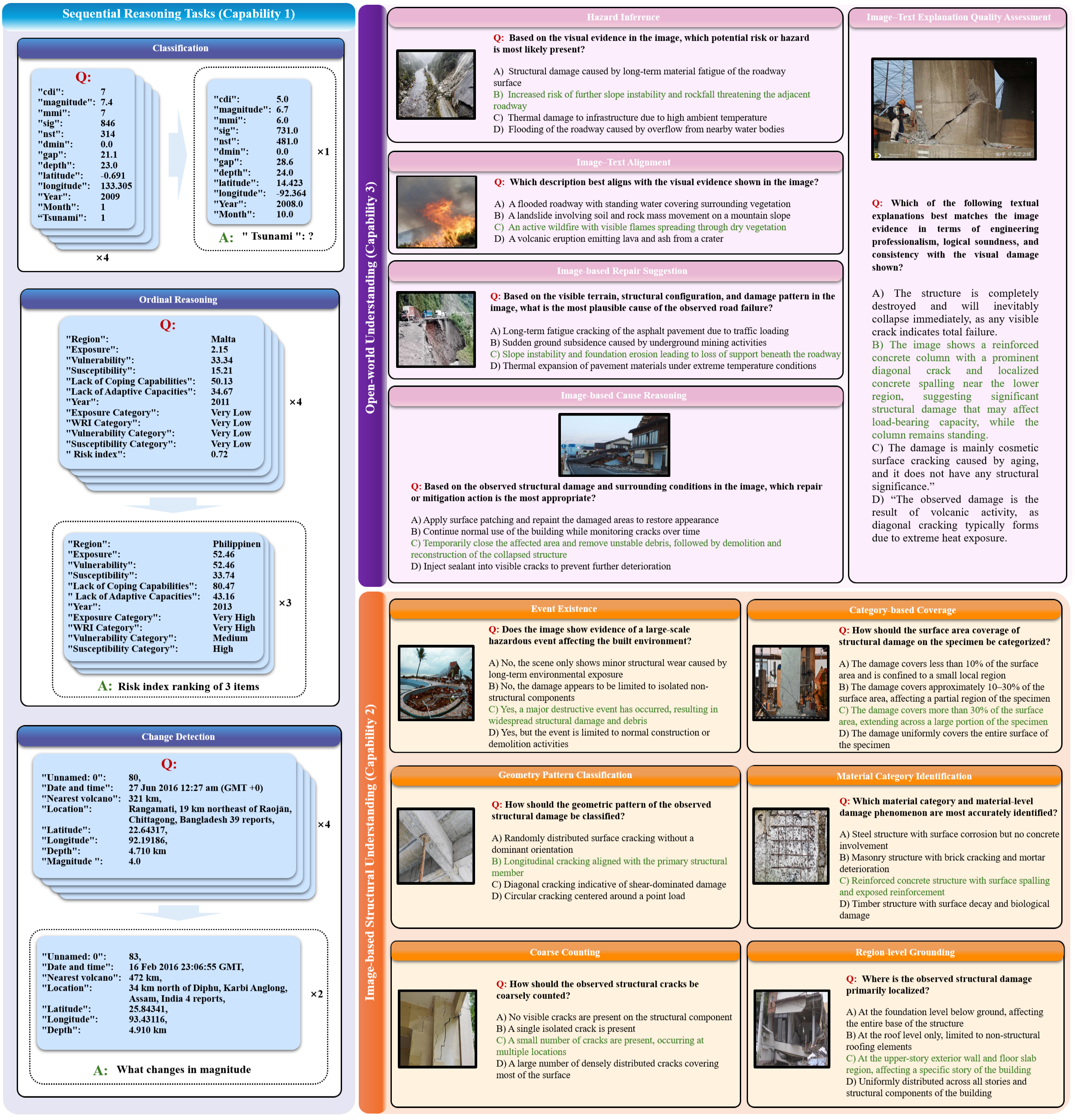}
	
	\caption{CivilBench task taxonomy and representative examples across all capability dimensions.}
	
	\label{fig2}
\end{figure*}

CivilBench is organized around a capability-oriented taxonomy that reflects the progression from structural perception to engineering reasoning and decision support, as illustrated in Figure~\ref{fig2}. The benchmark consists of three task categories, which are described below:
\begin{itemize}
    \item \textbf{Sequential Reasoning} evaluates temporal structural cognition by requiring models to infer future states, compare relative severity, or detect condition changes from limited historical observations. These tasks reflect degradation tracking and temporal consistency, which are essential for structural health monitoring.
    \item \textbf{Image-based Structural Understanding} assesses inspection capabilities from single images, including event detection, damage counting and coverage estimation, geometric pattern recognition, material identification, and region grounding. These tasks test whether models can extract reliable structural information from complex scenes.
    \item \textbf{Open-world Understanding} targets higher-level reasoning beyond direct perception. It includes hazard inference, cause analysis, explanation assessment, image-text alignment, and repair suggestion. These tasks require models to integrate visual evidence with engineering knowledge, making them closer to real-world diagnostic and decision-support scenarios.
\end{itemize}
These three categories provide a structured evaluation of VLMs across perception, temporal reasoning, domain knowledge, and practical engineering judgment.

\subsection{Evaluation Dimensions}
\label{sec34}

CivilBench contains 1,000 curated samples organized into three major categories and multiple fine-grained evaluation dimensions, as
summarized in Table~\ref{tab:capability_dimensions}. These dimensions are designed to diagnose model capability at
different levels of civil engineering intelligence, drawing on established evaluation tasks and engineering assessment practices.
Unless otherwise specified, the experimental evaluation is conducted on the valid subset after filtering.
We have the following capability dimensions:

\begin{itemize}
	\item \textbf{Sequential Reasoning}:
	Assesses temporal classification, severity comparison, and
	change detection to identify degradation trends and relative
	risks, drawing on established disaster damage assessment
	benchmarks~\citep{gupta2019creating}.
	
	\item \textbf{Image-Based Structural Understanding}:
	Evaluates visual inspection through event detection, damage
	counting and coverage estimation, geometry classification,
	material recognition, and damage localization, following
	established approaches to structural damage detection and
	quantification~\citep{zhou2023deep}.
	
	\item \textbf{Open-World Understanding}:
	Examines image--text consistency, hazard and cause reasoning,
	explanation quality, and repair or mitigation recommendations.
	These dimensions are motivated by prior studies on visual
	explanation assessment~\citep{he2026believing},
	damage cause inference~\citep{ccelik2026vision},
	and engineering repair recommendations
	~\citep{chou2026metaheuristic}.
\end{itemize}

Our taxonomy enables CivilBench to report both overall performance and fine-grained diagnostic profiles. Unlike generic VQA benchmarks focused on broad semantic understanding, CivilBench assesses domain-specific structural reasoning under engineering-relevant constraints.

\subsection{Dataset Statistics and Specific Features}

\begin{table*}[t]
	\centering
	\footnotesize
	\renewcommand{\arraystretch}{1.1}
	\caption{Capability-oriented evaluation dimensions in CivilBench.}
	\label{tab:capability_dimensions}
	
	\begin{tabular}{p{4.8cm} p{8.9cm}}
		\toprule
		\textbf{Dimension} & \textbf{Description} \\
		\midrule
		
		\multicolumn{2}{l}{\textbf{Capability 1: Sequential Reasoning Tasks}}\\
		\midrule
		
		Classification &
		Classify \colorbox{gray!20}{future event occurrence and states} from structured context. \\
		
		Ordinal Reasoning &
		Rank \colorbox{gray!20}{the severity levels} of observed damage or hazard events. \\
		
		Change Detection &
		Identify significant \colorbox{gray!20}{condition changes} over time. \\
		
		\midrule
		\multicolumn{2}{l}{\textbf{Capability 2: Image-based Structural Understanding}}\\
		\midrule
		
		Event Existence &
		Determine whether a \colorbox{gray!20}{hazardous structural event} is present. \\
		
		Coarse Counting &
		Estimate \colorbox{gray!20}{the quantity and spatial distribution} of damage. \\
		
		Category-based Coverage &
		Assess the \colorbox{gray!20}{coverage area} of visible damage. \\
		
		Geometry Pattern Classification &
		Classify \colorbox{gray!20}{damage patterns and dominant orientations} from visual cues. \\
		
		Material Category Identification &
		Identify \colorbox{gray!20}{material types and related damage} at the component level. \\
		
		Region-level Grounding &
		Localize \colorbox{gray!20}{structural damage regions} to specific components or zones. \\
		
		\midrule
		\multicolumn{2}{l}{\textbf{Capability 3: Open-world Understanding}}\\
		\midrule
		
		Image--Text Alignment &
		Evaluate \colorbox{gray!20}{semantic consistency} between image and text. \\
		
		Hazard Inference &
		Infer potential \colorbox{gray!20}{structural risks and hazards} from scene evidence. \\
		
		Explanation Quality Assessment &
		Assess the \colorbox{gray!20}{engineering quality} of explanations against visual evidence. \\
		
		Image-based Cause Reasoning &
		Reason about \colorbox{gray!20}{causes of observed damage} using contextual cues. \\
		
		Image-based Repair Suggestion &
		Propose reasonable \colorbox{gray!20}{repair or mitigation strategies} based on damage. \\
		
		\bottomrule
	\end{tabular}
\end{table*}


CivilBench is initially constructed with 1,000 carefully curated samples across Sequential Reasoning, Image-based Structural Understanding, and Open-world Understanding. Before evaluation, samples with invalid formatting, inconsistent annotations, missing assets, or processing failures are excluded to ensure reliable reporting. The final evaluated subset contains 972 valid samples: 285 Sequential Reasoning samples, consisting of 110 Classification, 90 Ordinal Reasoning, and 85 Change Detection samples; 286 Image-based Structural Understanding samples; and 401 Open-world Understanding samples. All experimental results and distribution-weighted Total scores are computed on this final evaluated subset. Image-based tasks adopt a single-choice MC-VQA format, where each sample contains an image, a question, four candidate answers, and one correct label. Sequential Reasoning Tasks use grouped pseudo-temporal records, enabling temporal classification, ordinal reasoning, and change detection under limited historical context.

CivilBench has three distinctive features. 1) CivilBench emphasizes domain-specific engineering semantics: questions require understanding of structural components, material conditions, damage mechanisms, and safety implications rather than generic object recognition. 2) CivilBench contains engineering-oriented visual content from inspection and disaster scenarios, where images often include occlusions, irregular geometries, complex backgrounds, and subtle damage cues. 3) CivilBench stresses decision-relevant reasoning, including hazard inference, failure-cause analysis, explanation assessment, and repair suggestion. These features make CivilBench a targeted benchmark for evaluating whether VLMs can support practical civil engineering understanding, rather than merely performing surface-level visual recognition.

\subsubsection{Data Distribution and Task Composition}

\begin{figure*}[t]
	\centering
	\includegraphics[width=\linewidth]{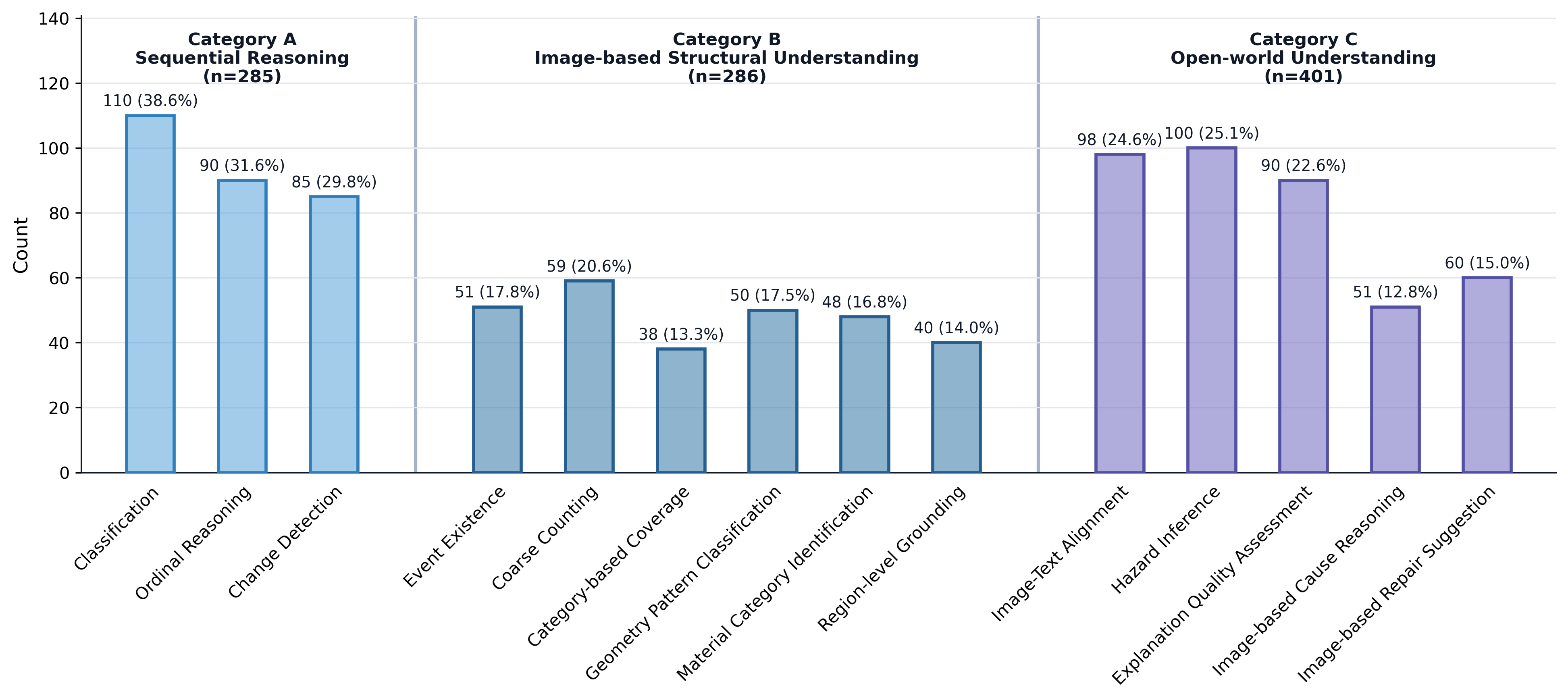}

	\caption{Sample distribution across CivilBench task categories and capability dimensions. Percentages are calculated within each category.}

	\label{fig:dataset_distribution}
\end{figure*}

To provide a clearer understanding of the dataset composition, we visualize the distribution of tasks and dimensions across the three major categories, as shown in Figure~\ref{fig:dataset_distribution}. For Category A (Sequential Reasoning Tasks), the dataset is relatively balanced across three sub-tasks, including classification (38.6\%), ordinal reasoning (31.6\%), and change detection (29.8\%). This balanced design ensures that the evaluation does not bias toward a particular reasoning type, while still covering diverse forms of structured inference, including label prediction, relative ordering, and pairwise comparison.

For Category B (Image-based Structural Understanding), coarse counting and event existence represent larger shares, reflecting the importance of detecting and quantifying structural damage. Geometry classification and material identification are also well represented, highlighting spatial and material cues. Region-level grounding and category-based coverage have fewer samples due to their precise annotation requirements and greater construction difficulty.

For Category C (Open-world Understanding), the dataset emphasizes high-level reasoning tasks, including hazard inference, image–text alignment, and explanation quality assessment, each with substantial samples. These tasks evaluate models' ability to integrate visual evidence with domain knowledge for decision-oriented reasoning. Repair suggestion and cause reasoning have moderate sample sizes, reflecting their higher annotation costs.

CivilBench has a deliberately non-uniform distribution that reflects each task's practical importance and annotation complexity, enabling realistic evaluation without over-representing simpler tasks.

\subsubsection{Image Resolution Distribution Analysis}

We further analyze the distribution of image resolutions in CivilBench. As shown in Figure~\ref{fig:resolution_distribution}, both Image-based Structural Understanding and Open-world Understanding cover a wide range of image resolutions, with most widths between approximately 300 and 1200 pixels. Open-world Understanding includes more high-resolution images, including samples exceeding 2000 pixels, reflecting its more complex scenes and richer visual details. Image-based Structural Understanding is more concentrated but still covers diverse visual conditions.

At the dimension level, \textit{Coarse Counting} and \textit{Event Existence} mainly use lower-resolution images, where global scene information is sufficient. Detail-oriented tasks, such as \textit{Geometry Pattern Classification}, \textit{Material Category Identification}, and \textit{Image-based Repair Suggestion}, generally require higher resolutions. Meanwhile, \textit{Hazard Inference} and \textit{Image-Text Explanation Quality Assessment} span a broader range, reflecting their reliance on both visual and contextual information. Overall, this diverse resolution distribution represents realistic engineering scenarios and reduces scale-specific overfitting.

\section{Experiments}
\label{Exp}

This section evaluates the performance of 17 representative VLMs on CivilBench. We first describe the experimental settings and benchmarking baselines, followed by the overall results and performance comparisons across different capability dimensions.
Finally, we further analyze the observed performance gaps and the main challenges faced by current VLMs.

\subsection{Experiment Settings}
We adopt a unified evaluation framework to ensure fair and reproducible comparison across different models, while allowing task-specific protocols tailored to distinct reasoning requirements. All models are evaluated using their official checkpoints without task-specific fine-tuning, and share identical input formats, prompt templates, parsing rules, and evaluation metrics.

\textbf{Evaluation metrics}. Across all tasks, we report accuracy as the primary evaluation metric, defined as the proportion of correctly predicted answers. In addition, we compute the average score (\textit{Avg}) across different task dimensions to reflect overall performance consistency, and the total score (\textit{Total}) as an aggregated indicator of model performance on the full benchmark. These metrics provide a concise yet informative assessment of model capability across diverse reasoning tasks.

\begin{center}
	\begin{minipage}{\columnwidth}
		\centering
		
		\includegraphics[width=0.85\linewidth]
		{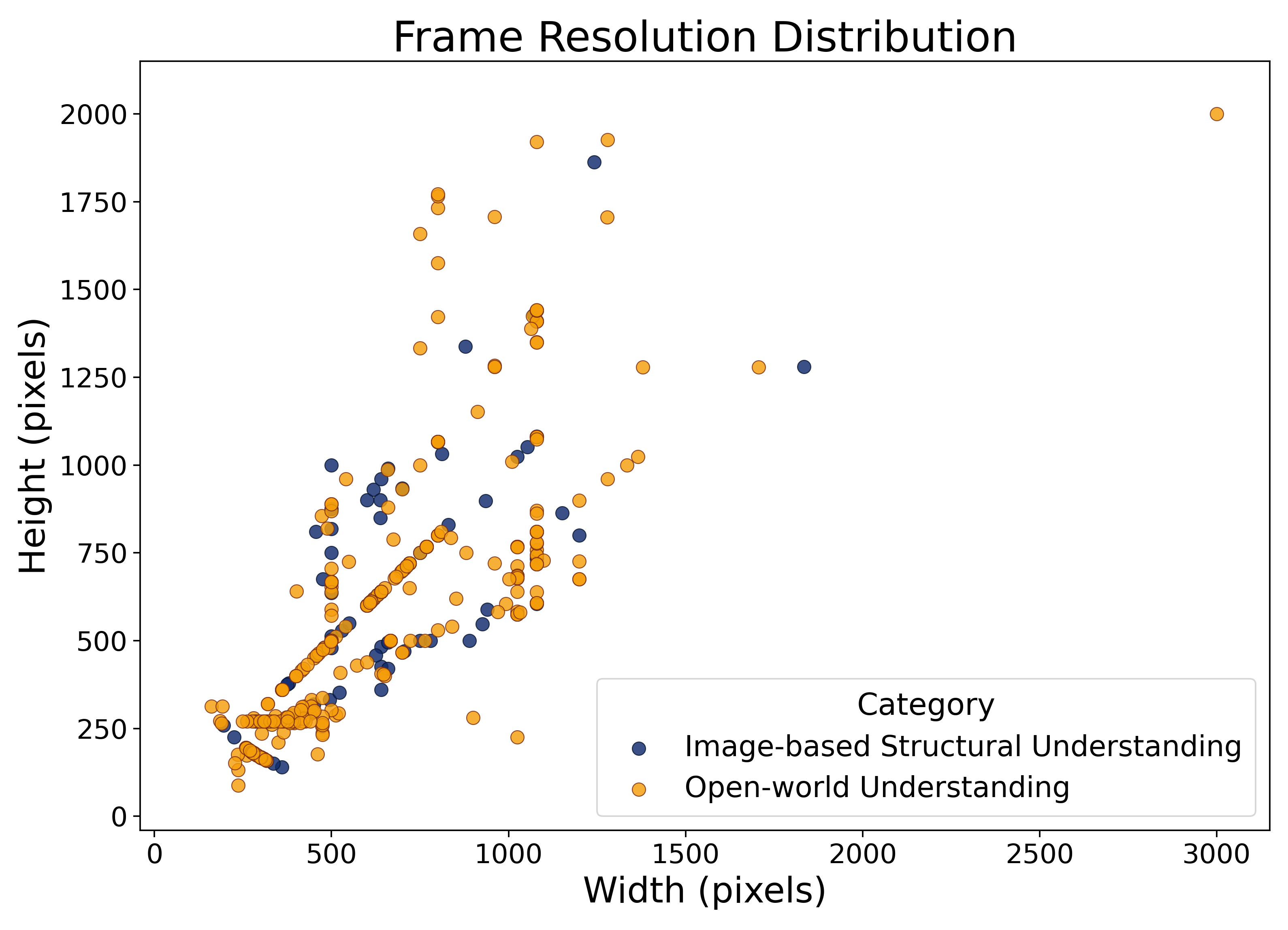}
		
		\small (a) Distribution by Category Image-based Structural Understanding and Open-world Understanding
		
		\vspace{0.8em}
		
		\includegraphics[width=0.85\linewidth]
		{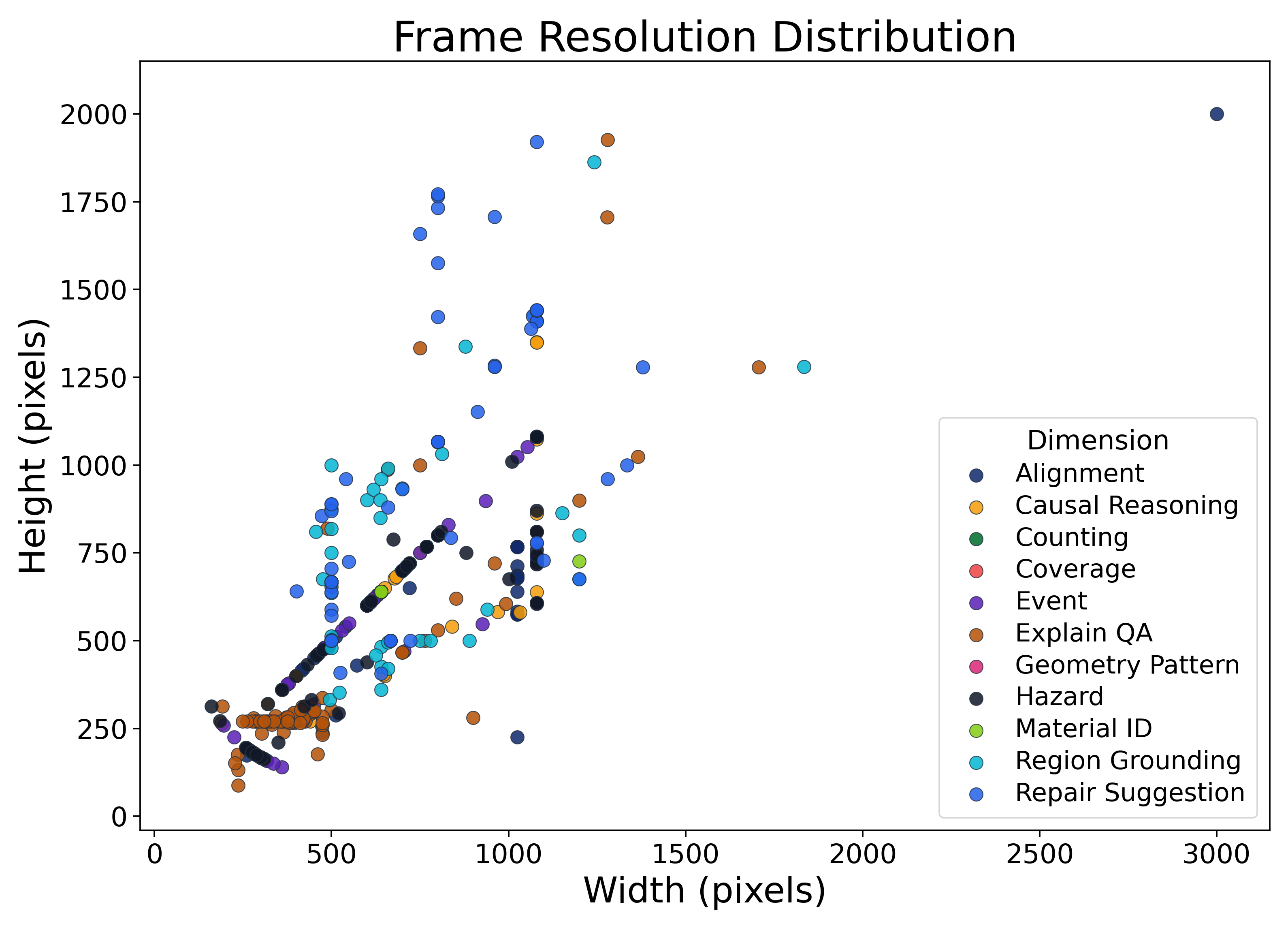}
		
		\small (b) Distribution by fine-grained dimensions

		\captionof{figure}{Image resolution distribution in CivilBench across (top) task categories and (bottom) fine-grained capability dimensions.}
		
		\label{fig:resolution_distribution}
		
	\end{minipage}
\end{center}

We evaluate Image-based Structural Understanding and Open-world Understanding using single-choice MC-VQA tasks. Each sample includes an image, a natural-language question, and four candidate answers. Models are instructed to return exactly one uppercase letter from \{A, B, C, D\}, without explanation. A rule-based parser extracts predictions, while incorrectly formatted outputs are marked as invalid.

\textbf{Sequential Reasoning tasks}. For Sequential reasoning tasks, we adopt a few-shot in-context learning setting, where each sample is formulated as a structured reasoning problem derived from tabular or statistical data. Each instance consists of a set of context examples (input--label pairs) and a query set, and models are required to infer the underlying mapping rule before making predictions on unseen inputs. This setting evaluates the ability of models to perform structured reasoning under limited observations.

We consider three types of sequential reasoning tasks:

\textbf{1) Classification.} Models are provided with several input--label examples and are required to predict the label of a test input. The label space is explicitly defined in the prompt, and models must output a single valid label.

\textbf{2) Ordinal reasoning.} Models are required to infer latent scoring rules from context examples and identify the relative ordering among multiple candidates. Specifically, models must predict both the maximum and minimum indices within a candidate set. Outputs are constrained to structured formats (\emph{e.g.}, \texttt{MAX\_INDEX} and \texttt{MIN\_INDEX}), and only valid index predictions are accepted.

\textbf{3) Change detection.} Models are asked to compare candidates and determine which one reflects a more severe condition based on inferred patterns. Similar to ordinal reasoning, predictions are restricted to local indices and validated through rule-based parsing.

To reduce stochastic variability and ensure consistent outputs, generation length is restricted, and decoding settings are fixed across all models.

\subsection{Benchmarking Baselines}

We evaluate a diverse set of representative VLMs, including both open-source and closed-source models with different architectures, parameter scales, and training strategies.

\textbf{Open Source VLMs.}
We include several widely used open-source models, including MiniCPM-V~\citep{yao2024minicpm}, a lightweight multimodal model designed for efficient visual reasoning; Qwen2.5-VL-7B and Qwen2.5-VL-3B-Instruct, which are part of the Qwen-VL series with strong multimodal alignment capabilities~\citep{bai2023qwen}; InternVL and its variants (InternVL3-1B and InternVL3\_5-14B), which focus on scaling vision-language pretraining for improved perception and reasoning~\citep{chen2024internvl}; LLaVA-OneVision-7B, a unified multimodal model based on the LLaVA framework~\citep{liu2023visual}; and Phi-3.5-Vision, a compact vision-language model with competitive performance at small scale~\citep{abdin2024phi}. 
We also include Qwen3-VL-8B-Instruct~\citep{bai2025qwen3}, a more recent version with improved reasoning ability, as well as DeepSeek-VL2~\citep{wu2024deepseek}, which emphasizes multimodal reasoning and instruction-following capabilities. 
In addition, Mistral-Small-3.1-24B-Instruct-2503~\citep{siino2024prompt} is evaluated as a strong open-source LLM baseline with extended reasoning capacity.

\textbf{Closed Source VLMs.}
We further include several closed-source VLMs with strong multimodal capabilities, including Qwen-VL-Plus, a high-performance commercial variant of the Qwen-VL series~\citep{wu2025qwen}, GPT-4.1-mini, a lightweight yet capable multimodal model, GPT-4.1, which serves as a strong baseline with advanced reasoning and multimodal understanding capabilities~\citep{openai2025gpt41}, GPT-5-mini, an efficient reasoning-oriented multimodal model with improved generalization and visual reasoning ability, GPT-5, the latest flagship multimodal model from OpenAI~\citep{singh2025openai}, and Gemini-2.5-Pro~\citep{comanici2025gemini}, Google's flagship multimodal model with advanced visual understanding, long-context reasoning, and strong performance across diverse multimodal tasks.

These models together provide comprehensive coverage of different model families, scales, and design paradigms, enabling a systematic comparison of their performance on CivilBench.

\subsection{Results and Observations}

\begin{table*}[t]
\centering
\caption{Performance comparison of different VLMs on CivilBench (\%). The best result in each column is highlighted in bold. A1: Classification, A2: Ordinal Reasoning, and A3: Change Detection, respectively; B: Image-based Structural Understanding; C: Open-world Understanding. ``Avg'' represents the capability-balanced average accuracy across all evaluated dimensions, whereas ``Total'' represents the distribution-weighted overall accuracy.}
\label{tab:main_results}
\resizebox{\textwidth}{!}{
\begin{tabular}{lccccccc}
\toprule
\textbf{Model} & \textbf{A1} & \textbf{A2} & \textbf{A3} & \textbf{B} & \textbf{C} & \textbf{Avg} & \textbf{Total} \\
\midrule
\multicolumn{8}{c}{\textbf{Open Source}} \\
\midrule
Minicpm-V~\citep{yao2024minicpm} & 64.55 & 20.00 & 43.53 & 43.36 & 42.14 & 42.72 & 43.13 \\
Qwen2.5-VL-7B~\citep{bai2023qwen} & 47.27 & 26.67 & 54.12 & 47.90 & \textbf{88.28} & 52.85 & 63.07 \\
InternVL~\citep{chen2024internvl} & 65.45 & 26.67 & 61.18 & 50.70 & 84.79 & 57.76 & 65.13 \\
LLaVA-OneVision-7B~\citep{liu2023visual} & 70.00 & 13.33 & 49.41 & 45.45 & 84.54 & 52.55 & 61.73 \\
Phi-3.5-Vision~\citep{abdin2024phi} & 64.55 & 11.11 & 43.53 & 39.51 & 85.04 & 48.75 & 58.85 \\
Qwen2.5-VL-3B-Instruct~\citep{bai2023qwen} & 4.55 & 21.11 & 10.59 & 46.15 & 84.79 & 33.44 & 51.95 \\
InternVL3-1B~\citep{chen2024internvl} & 60.00 & 1.11 & 49.41 & 37.06 & 80.55 & 45.63 & 55.33 \\
Qwen3-VL-8B-Instruct~\citep{bai2025qwen3} & 70.00 & 25.56 & 57.65 & 52.45 & 86.28 & 58.39 & 66.37 \\
DeepSeek VL2~\citep{wu2024deepseek} & 66.36 & 25.56 & 45.88 & 36.01 & 59.85 & 46.73 & 49.27 \\
InternVL3\_5-14B-HF~\citep{chen2024internvl} & 70.00 & 30.00 & 43.53 & 47.55 & 85.79 & 55.37 & 63.88 \\
Mistral-Small-3.1-24B-Instruct-2503~\citep{siino2024prompt} & 69.09 & 27.78 & 57.65 & 48.60 & 87.78 & 58.18 & 65.95 \\
\midrule
\multicolumn{8}{c}{\textbf{Closed Source}} \\
\midrule
Qwen-VL-Plus~\citep{wu2025qwen} & 70.91 & 28.89 & 57.65 & 53.15 & 85.54 & 59.23 & 66.67 \\
GPT-4.1-mini~\citep{openai2025gpt41} & \textbf{74.55} & 25.56 & 54.12 & 43.01 & 81.80 & 55.81 & 61.94 \\
GPT-4.1~\citep{openai2025gpt41} & 69.09 & 27.78 & 43.53 & 59.09 & 75.56 & 55.01 & 62.76 \\
GPT-5-mini~\citep{singh2025openai} & 70.00 & 26.67 & 62.35 & 52.10 & 82.29 & 58.68 & 65.12 \\
GPT-5~\citep{singh2025openai} & 73.64 & 28.89 & \textbf{67.06} & 55.59 & \textbf{88.28} & 62.69 & 69.65 \\
Gemini-2.5-pro~\citep{comanici2025gemini} & 70.00 & \textbf{37.78} & 63.49 & \textbf{60.49} & 88.03 & \textbf{63.96} & \textbf{71.09} \\
\bottomrule
\end{tabular}
}
\end{table*}

Table~\ref{tab:main_results} reports the performance of representative vision-language models on CivilBench across different task categories, including Sequential Reasoning Tasks (A1--A3), Image-based Structural Understanding (B), and Open-world Understanding (C), together with the capability-balanced average accuracy (Avg) and the distribution-weighted total accuracy (Total). 
Figure~\ref{fig:CivilBench_diagnostic_avg_total} further compares the capability-balanced Avg score with the distribution-weighted Total score across all evaluated models. The gap between the two scores is mainly driven by the larger Image-based Structural Understanding and Open-world Understanding subsets, whose sample sizes give them greater influence on the distribution-weighted Total score.

Overall, Avg and Total scores are positively correlated but capture distinct performance aspects. Total scores are generally higher because larger, relatively easier image-based and open-world categories dominate the evaluated subset, potentially masking weaknesses in smaller, reasoning-intensive dimensions. The color coding in Figure~\ref{fig:CivilBench_diagnostic_avg_total} makes this effect more explicit.  Models in the upper-right region perform strongly overall but remain limited in Ordinal Reasoning, showing that high aggregate accuracy does not imply balanced civil-engineering reasoning. Gains often stem from visual recognition, language priors, and open-world knowledge, while ordinal reasoning remains a bottleneck. Thus, Avg reveals cross-dimensional imbalance, whereas Total reflects performance under CivilBench's sample distribution.
To support the following capability-level discussion, Table~\ref{tab:capability_numeric_diagnostic} further summarizes the results across dimensions, including sample share, mean accuracy, score range, open--closed model gap, and the number of models falling below or above key accuracy thresholds.

\textbf{Performance discrepancy across task categories.}
Open-world Understanding is the strongest category for most models, with an average accuracy of 80.67\%; 14 out of 17 models score above 80\%. 
By contrast, Ordinal Reasoning is consistently difficult. 
Its average accuracy is only 23.79\%, and even the best model, Gemini-2.5-pro, reaches only 37.78\%. 
Most models remain below 30\% on this dimension. 
The gap suggests that current VLMs are much better at answering visually grounded or knowledge-aligned questions than at recovering relative numerical relations from a few examples.

\textbf{Sequential Reasoning as a bottleneck.}
The three sequential reasoning tasks show different difficulty levels. 
Classification is relatively easier, with an average accuracy of 63.53\% and a best score of 74.55\% by GPT-4.1-mini. 
Change Detection is harder, averaging 50.86\%, with GPT-5 reaching 67.06\%. 
Ordinal Reasoning is the clear bottleneck. 
Its best score is still lower than the average accuracy of most other dimensions. 
This means the main challenge is not simply reading sequential inputs, but inferring an ordering rule from very limited observations.

\begin{center}
	\begin{minipage}{\columnwidth}
		\centering
		
		\includegraphics[width=\linewidth]{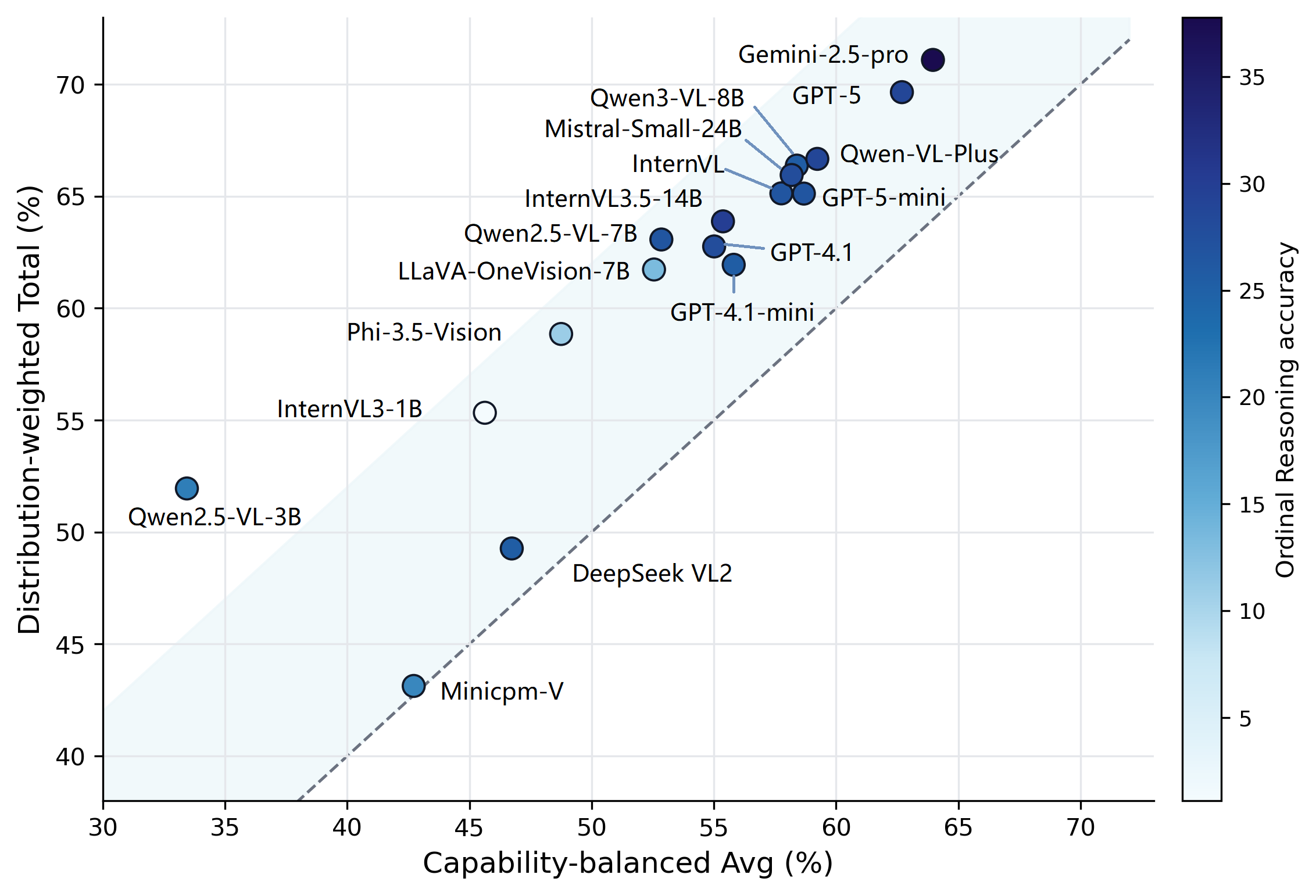}

		\captionof{figure}{Comparison of capability-balanced Avg and distribution-weighted Total accuracy across VLMs on CivilBench. Color indicates Ordinal Reasoning accuracy.}
		
		\label{fig:CivilBench_diagnostic_avg_total}
		
	\end{minipage}
\end{center}

\textbf{Moderate performance in structural understanding.}
Image-based Structural Understanding falls between sequential reasoning and open-world tasks. 
The scores range from 36.01\% to 60.49\%, with an average of 48.13\%. 
Gemini-2.5-pro performs best in this category, while GPT-4.1 is close behind with 59.09\%. 
These numbers show that models can extract some structural cues from images, but still struggle with fine-grained damage patterns, material categories, spatial grounding, and coverage estimation.

\textbf{High accuracy in Open-world tasks.}
Open-world Understanding is much easier for current models. 
Qwen2.5-VL-7B and GPT-5 both reach 88.28\%, and several other models exceed 85\%. However, a high Open-world Understanding task score does not always mean balanced performance. For example, Qwen2.5-VL-3B-Instruct obtains 84.79\% on Open-world Understanding, but its Avg is only 33.44\% because it performs poorly on the Sequential Reasoning tasks. This pattern suggests that some open-world questions are more aligned with language priors or pretrained knowledge than with the structured reasoning required by Sequential Reasoning tasks.

\textbf{Importance of distribution-aware evaluation.}
``Avg'' and ``Total'' provide complementary performance views. 
Avg weights the five capability dimensions equally, whereas Total reflects CivilBench's sample distribution.
As shown by the Share column in Table~\ref{tab:capability_numeric_diagnostic}, Image-based Structural Understanding and Open-world Understanding account for 29.42\% and 41.26\% of the final evaluated subset, respectively.
The mean Total score (61.29\%) exceeds the mean Avg score (53.40\%), largely because the larger Image-based Structural Understanding and Open-world Understanding subsets contribute more to Total.
Consequently, high Total scores may mask weaknesses in Ordinal Reasoning. Reporting both metrics therefore captures overall benchmark performance while revealing capability imbalances.

\textbf{Overall observation.}
Gemini-2.5-pro achieves the best overall results, with an Avg of 63.96\% and a Total of 71.09\%. 
Still, its Ordinal Reasoning accuracy is only 37.78\%. 
This shows that stronger general-purpose VLMs have not yet solved sparse numerical reasoning and comparison-based inference. 
CivilBench reveals a clear gap between models' performance on perception-oriented or knowledge-aligned tasks and their ability to conduct structured reasoning for engineering analysis.

\begin{table*}[t]
	\centering
	\footnotesize
	\renewcommand{\arraystretch}{1.12}
	
	\caption{Capability-level performance statistics across all evaluated models on CivilBench. A1--A3 denote Classification, Ordinal Reasoning, and Change Detection; B and C denote Image-based Structural Understanding and Open-world Understanding. Share indicates the proportion of evaluated samples; Mean, Std, Min, Max, and Range summarize model accuracy. Open and Closed report mean accuracy by model type, with $\Delta$ indicating their gap. The final two columns count models with accuracy below 30\% and at least 80\%. Bold values highlight key diagnostic findings.}
	
	\label{tab:capability_numeric_diagnostic}
	\resizebox{\textwidth}{!}{
		\begin{tabular}{lcccccccccccc}
			\toprule
			\textbf{Dimension} & \textbf{$n$} & \textbf{Share} & \textbf{Mean} & \textbf{Std} & \textbf{Min} & \textbf{Max} & \textbf{Range} & \textbf{Open} & \textbf{Closed} & \textbf{$\Delta$} & \textbf{\# $<30$} & \textbf{\# $\geq 80$} \\
			\midrule
			A1 & 110 & 11.32\% & 63.53 & 15.94 & 4.55 & 74.55 & 70.00 & 59.26 & 71.36 & 12.11 & 1  & 0  \\
			A2 & 90  & 9.26\%  & \textbf{23.79} & 8.26  & 1.11 & \textbf{37.78} & 36.67 & 20.81 & 29.26 & 8.45  & \textbf{15} & 0  \\
			A3 & 85  & 8.74\%  & 50.86 & 12.57 & 10.59 & 67.06 & 56.47 & 46.95 & 58.03 & 11.08 & 1  & 0  \\
			B  & 286 & 29.42\% & 48.13 & 6.86  & 36.01 & 60.49 & 24.48 & 44.98 & 53.91 & 8.93  & 0  & 0  \\
			C  & 401 & \textbf{41.26\%} & \textbf{80.67} & 11.68 & 42.14 & 88.28 & 46.14 & 79.08 & 83.58 & 4.51  & 0  & \textbf{14} \\
			\bottomrule
		\end{tabular}
	}
\end{table*}

\subsection{Further Analysis}

In this section, we further analyze model behavior on CivilBench and discuss what the results reveal about current VLMs in civil-engineering scenarios.

\textbf{Capability-dependent difficulty hierarchy.}
Table~\ref{tab:main_results} shows a clear capability-dependent difficulty hierarchy.  Open-world Understanding tasks achieve the strongest overall performance, while Ordinal Reasoning tasks are consistently the most challenging dimension. Across all evaluated models, the average accuracy on Ordinal Reasoning tasks is only about $23.8\%$, and even the best model reaches only $37.78\%$. This is in sharp contrast to Open-world Understanding tasks, whose average accuracy is about $80.7\%$ and where several models exceed $85\%$ accuracy.

This gap suggests that current VLMs are not uniformly weak on civil-engineering tasks. Instead, their failures are concentrated in tasks that require structured comparison, ordinal judgment, and stage-wise reasoning. 
Open-world Understanding tasks can often be solved by combining visual recognition with broad world knowledge and language priors. 
In contrast, Ordinal Reasoning tasks require models to compare multiple observations and infer relative severity or risk levels, which cannot be reliably solved by recognizing isolated visual or textual cues.

\begin{figure*}[t]
\centering
\includegraphics[width=0.80\textwidth]{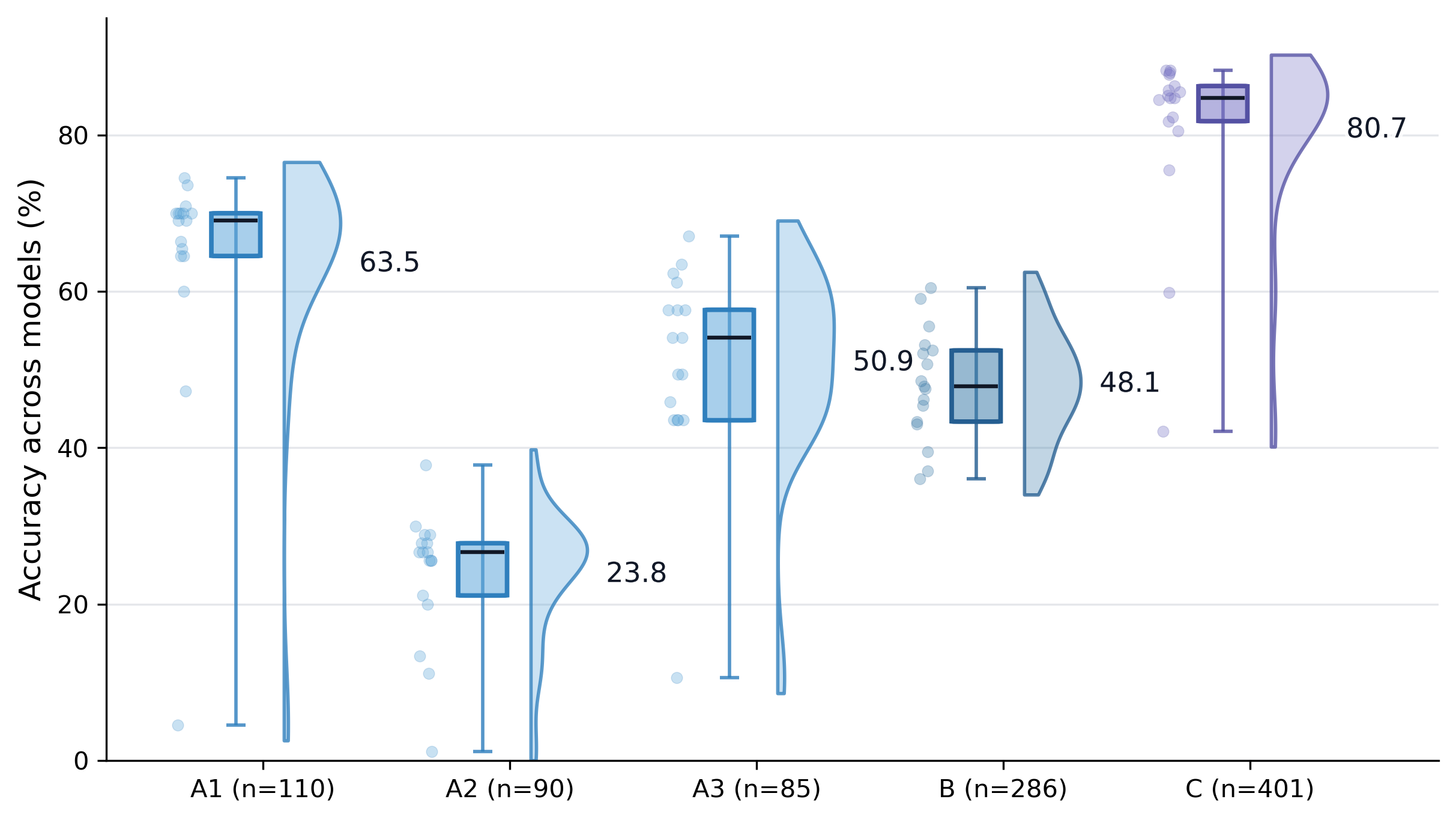}
\caption{Model accuracy distributions across CivilBench capability dimensions. Raincloud plots show individual model accuracies, boxplots, and density distributions; annotated values indicate mean accuracy, and $n$ denotes sample size.}

\label{fig:CivilBench_diagnostic_difficulty}
\end{figure*}

Figure~\ref{fig:CivilBench_diagnostic_difficulty} highlights clear difficulty differences across capability dimensions, showing model accuracy distributions and sample sizes. Ordinal Reasoning has the lowest mean accuracy and upper performance bound despite not being the smallest subset. Its best-performing model, Gemini-2.5-pro, reaches only 37.78\%, indicating an intrinsic difficulty in inferring ordinal relations from sparse structured evidence rather than an effect of sample imbalance. In contrast, Open-world Understanding has the largest sample size and highest average performance, with Qwen2.5-VL-7B and GPT-5 jointly achieving 88.28\%, suggesting stronger alignment with pretrained knowledge and language priors. The leading model also varies by dimension: GPT-4.1-mini leads Classification, Gemini-2.5-pro leads Ordinal Reasoning and Image-based Structural Understanding, GPT-5 leads Change Detection, and Qwen2.5-VL-7B and GPT-5 tie in Open-world Understanding. These varying rankings show that aggregate scores can obscure capability-specific strengths and weaknesses.

\textbf{Why are Ordinal Reasoning and Change Detection difficult?} As summarized in Table~\ref{tab:a2_a3_regression_target_burden}, these tasks require models to infer an unknown ranking or change rule from only four labeled examples and then apply it to a new instance. The benchmark draws Ordinal Reasoning from 10 source schemas and Change Detection from 17, while each instance contains an average of 8.8 and 15.9 unique fields, respectively. Models must therefore identify the fields relevant to the target relation with minimal supervision. This challenge differs from standard classification because 88.9\% of Ordinal Reasoning labels and 94.1\% of Change Detection labels are numeric, requiring models to derive implicit relations rather than recognize predefined categories. Noisy or partially observable relationships and repeated labels further reduce the information available for distinguishing fine-grained ordinal or temporal.

In addition to this regression-derived target burden, some samples also involve incomplete or knowledge-heavy evidence, as illustrated in Figure~\ref{fig:a2_a3_difficulty_decomposition}. 
Here, event-record cues refer to explicit mentions of real disasters or infrastructure events, missing cells indicate incomplete structured records, and long text fields introduce additional information-filtering demands.

For such cases, correct prediction may require more than local tabular evidence: models may need to combine sparse numerical reasoning with civil-engineering domain knowledge, parametric knowledge acquired during pretraining, or external historical context. Therefore, the difficulty of these two dimensions reflects a joint challenge of data-efficient numerical rule induction, robust relational comparison, and knowledge-grounded decision making.

\begin{center}
	\begin{minipage}{\columnwidth}
		\centering
		
		\includegraphics[width=\linewidth]{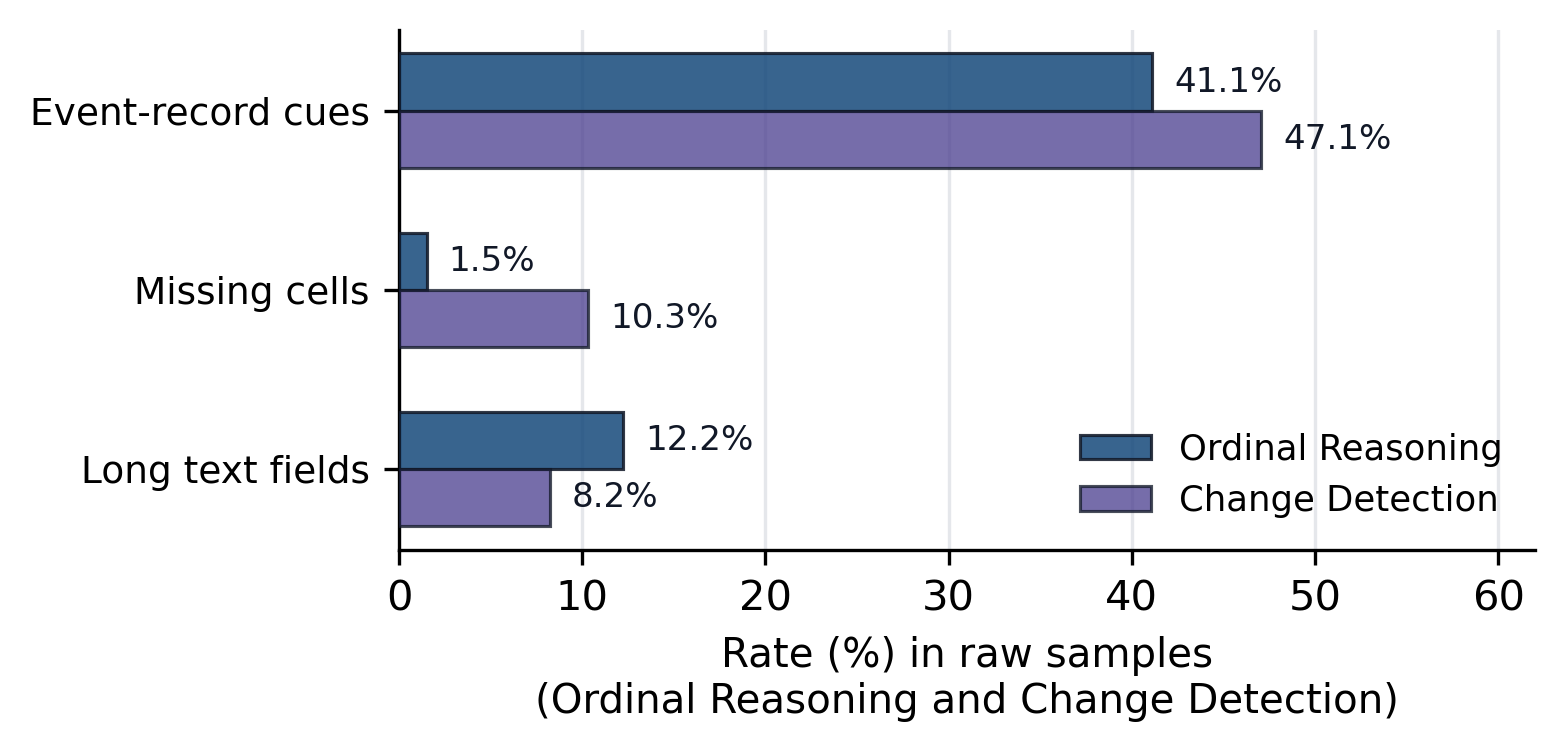}

		\captionof{figure}{Distribution of event-record cues, missing cells, and long text fields in raw Ordinal Reasoning and Change Detection samples.}
		\label{fig:a2_a3_difficulty_decomposition}

	\end{minipage}
\end{center}

\begin{table*}[t]
\centering
\footnotesize
\renewcommand{\arraystretch}{1.12}
\caption{Regression-derived target burden in Ordinal Reasoning and Change Detection. 
Statistics are computed from the final evaluated Category A samples, with 90 Ordinal Reasoning samples and 85 Change Detection samples. 
Values in parentheses denote ranges, and values with $\pm$ denote the mean and standard deviation. 
The ``fields per labeled observation'' value measures how many structured input attributes are associated with each support label, reflecting the variable-selection burden under sparse evidence.}
\label{tab:a2_a3_regression_target_burden}
\begin{tabular}{m{4.5cm}ccm{5.3cm}}
\toprule
\textbf{Statistic} & \textbf{Ordinal Reasoning} & \textbf{Change Detection} & \textbf{Interpretation} \\
\midrule
Source schemas & 10 & 17 & Instances come from \colorbox{gray!20}{regression sources}, not one rule. \\
\specialrule{0.3pt}{0pt}{0pt}
Samples per source schema & 9.0 (5--10) & 5.0 (5--5) & Each source relation contributes only a \colorbox{gray!20}{small number} of benchmark instances. \\
\specialrule{0.3pt}{0pt}{0.2pt}
Labeled observations per instance & 4 (4--4) & 4 (4--4) & Only four \colorbox{gray!20}{support examples} are available for inferring the hidden target relation. \\
\specialrule{0.3pt}{0pt}{0pt}
Query observations per instance & 3 (3--3) & 2 (2--2) & The inferred rule must be applied to \colorbox{gray!20}{unseen candidates} or a held-out pair. \\
\specialrule{0.3pt}{0pt}{0pt}
Unique structured fields per instance & 8.8 $\pm$ 3.7 & 15.9 $\pm$ 11.9 & The model must select \colorbox{gray!20}{relevant variables} from heterogeneous structured inputs. \\
\specialrule{0.3pt}{0pt}{0pt}
Fields per labeled observation & 2.21 $\pm$ 0.93 & 3.97 $\pm$ 2.99 & Larger value indicates \colorbox{gray!20}{candidate variables} for each observed target label. \\
\specialrule{0.3pt}{0pt}{0pt}
Numeric-label sequences & 88.9\% & 94.1\% & Most targets are \colorbox{gray!20}{numeric}, reflecting their regression-derived origin. \\
\specialrule{0.3pt}{0pt}{0pt}
Repeated-label sequences & 26.7\% & 32.9\% & Tied support labels reduce the observable \colorbox{gray!20}{ordering or change signal}. \\
\bottomrule
\end{tabular}
\end{table*}

\textbf{Distribution-aware evaluation.}

CivilBench also highlights the importance of distinguishing between distribution-weighted and capability-balanced evaluation. Let $n_k$ denote the number of samples in capability dimension $k$, and let $N$ be the total number of samples. The Total score can be interpreted as:
\begin{equation}
	\mathrm{Total}
	=
	\sum_{k}
	\frac{n_k}{N}
	\mathrm{Acc}_k,
	\label{eq:total_score}
\end{equation}
whereas Avg assigns equal weight to each evaluated capability:
\begin{equation}
	\mathrm{Avg}
	=
	\frac{1}{K}
	\sum_{k}
	\mathrm{Acc}_k.
	\label{eq:avg_score}
\end{equation}
This distinction is important because the evaluated subset is not uniformly distributed across dimensions.  Image-based Structural Understanding and Open-world Understanding contain 286 and 401 samples, respectively, accounting for 70.7\% of the final evaluated subset, whereas Ordinal Reasoning contains only 90 samples.  As a result, Total accuracy is more strongly affected by performance on the larger perception-driven and open-world categories. 
For example, Qwen2.5-VL-3B-Instruct obtains a low capability-balanced Avg of 33.44\%, but its Total score rises to 51.95\%, mainly because it performs much better on Image-based Structural Understanding and Open-world Understanding than on the sequential reasoning dimensions. 
Similarly, InternVL3-1B and DeepSeek VL2 have comparable Avg scores, 45.63\% and 46.73\%, but their Total scores differ noticeably, 55.33\% versus 49.27\%, largely due to the much higher Open-world Understanding accuracy of InternVL3-1B. 

Figure~\ref{fig:CivilBench_diagnostic_avg_total} further illustrates that models with similar Avg scores may obtain different Total scores because their strengths are distributed differently across capability dimensions. In particular, strong performance on Open-world Understanding can improve Total accuracy, but it does not necessarily indicate robust sequential reasoning. 
For instance, Qwen2.5-VL-7B reaches 88.28\% on Open-world Understanding and 63.07\% in Total accuracy, while its Ordinal Reasoning accuracy remains only 26.67\%. Therefore, the gap between Avg and Total provides a useful diagnostic signal: it reveals whether a model is broadly balanced or mainly benefits from high performance on the dominant task categories.

\textbf{Methodological implications.} These findings indicate that improving CivilBench performance requires more than stronger visual perception or large-scale instruction tuning. Sequential Reasoning may benefit from table-aware reasoning, temporal state tracking, ordinal ranking, pairwise comparison, and cross-observation consistency checking, supported by ranking-based supervision, preference modeling, programmatic numerical reasoning, and uncertainty-aware prediction rather than standard multiple-choice classification alone. Retrieval augmentation may also be needed for historical disasters and region-specific infrastructure events when relevant evidence is absent from the prompt. For Image-based Structural Understanding, models require explicit structural representations, including region-level grounding, component-aware features, damage localization, and geometry-sensitive pretraining, because engineering decisions depend on where damage occurs, which components are affected, and how damage patterns relate to material and structural behavior.

\subsection{Enhancing Numerical Reasoning with Priors and Web Retrieval}
\label{sec:a2_a3_enhancement}

The preceding analysis reveals a key limitation of current vision-language models in Ordinal Reasoning (A2) and Change Detection (A3): they struggle to reliably identify numerical relationships from sparse labeled observations, particularly when relevant event-specific knowledge is absent from the input. This suggests that their numerical reasoning remains sensitive to the availability of explicit guidance and external evidence. To further investigate this limitation, we conduct supplementary experiments with GPT-5-mini using two configurations: \textit{prior-guided inference} and \textit{retrieval-augmented prior-guided inference}. These experiments examine whether explicit numerical priors and access to external knowledge can alleviate the observed difficulties without modifying the benchmark instances or increasing the number of labeled support examples.

\textbf{Experimental setup.}
Both configurations are evaluated on the original 90 A2 and 85 A3 instances, each with four labeled support observations. All benchmark inputs and annotations remain unchanged, and predictions are evaluated using the original parser and accuracy metric, with invalid outputs counted as incorrect. The main-experiment GPT-5-mini scores (26.67\% for A2 and 62.35\% for A3) serve as the baseline. As the configurations were evaluated in separate runs, performance differences are reported descriptively without claims of statistical significance or isolated causal effects.

\textbf{Prior-guided inference.}
The original prompt asks the model to infer a hidden numerical relation from
four examples. Prior-guided inference additionally identifies the target indicator, such as
earthquake magnitude, event duration, or estimated financial loss, and specifies
units when these are established by the available information.
It also provides schema-specific numerical priors. Two examples relevant to
disaster-risk assessment and event-duration analysis are:
\begin{equation}
	\begin{aligned}
		R_i^{\mathrm{WRI}} &= \frac{E_i V_i}{100},\\
		T_i^{\mathrm{event}} &= \bigl(t_i^{\mathrm{end}}-t_i^{\mathrm{start}}\bigr)_{\mathrm{days}}.
	\end{aligned}
	\label{eq:a2_a3_formula_priors}
\end{equation}
Here, $E_i$ and $V_i$ denote the exposure and vulnerability percentages of record $i$, respectively, and $R_i^{\mathrm{WRI}}$ represents its World Risk Index. $T_i^{\mathrm{event}}$ denotes the event duration in days, calculated from the recorded start and end dates. For the GDP schema, the product of per-capita GDP and total population is also considered as an arithmetic prior.
These priors are supplied only for the corresponding target schemas and only
when they agree with the current instance's support labels under predefined
tolerances. No query target values are supplied or calculated for inclusion
in the prompt. Additional instructions discourage unsupported deterministic
rules, distinguish missing values from zero, and require consistent comparison
across candidates. For A3, the prompt explicitly follows the implemented
evaluation objective: selecting the candidate with the larger target value.

\textbf{Retrieval-augmented inference.}
Retrieval-augmented inference extends Prior-guided inference by introducing a web-retrieval stage before prediction. The retrieval process uses available event names, dates, locations, coordinates, identifiers, and record URLs to search for relevant historical records or indicator definitions. The model is instructed to prioritize primary sources and assess event identity, temporal coverage, and indicator definitions before incorporating retrieved information.

The retrieved evidence is summarized and appended to the prediction prompt. When no reliable match is identified, the model is instructed to rely on the labeled support examples and applicable numerical priors. The benchmark's ground-truth annotations are not explicitly provided to either stage. However, public records retrieved during this process may contain the corresponding target values, as the experiment evaluates an open-retrieval setting.
The prediction stage retains the inference settings of Prior-guided inference, allowing the additional effect of retrieved evidence to be examined under the same prediction configuration. All 175 instances receive a completed retrieval response.

\begin{table*}[t]
	\centering
	\small
	\renewcommand{\arraystretch}{1.15}
	\caption{Performance comparison of GPT-5-mini on A2 and A3 under different inference configurations. Accuracy is reported in percent (\%), and improvements over the baseline are given in percentage points (pp). The retrieval-augmented configuration extends prior-guided inference with web retrieval.}
	\label{tab:a2_a3_enhancement}
	\begin{tabular}{lrrrr}
		\toprule
		\textbf{Configuration} & \textbf{A2} & \textbf{A3}
		& \textbf{$\Delta$ A2 (pp)} & \textbf{$\Delta$ A3 (pp)} \\
		\midrule
		Reported baseline & 26.67 & 62.35 & --- & --- \\
		Prior-guided inference & 35.56 & 68.24 & +8.89 & +5.89 \\
		Prior-guided inference + retrieval & \textbf{46.67} & \textbf{71.76}
		& \textbf{+20.00} & \textbf{+9.41} \\
		\bottomrule
	\end{tabular}
\end{table*}

\textbf{Results and interpretation.}
As shown in Table~\ref{tab:a2_a3_enhancement}, prior-guided inference improves GPT-5-mini's accuracy from 26.67\% to 35.56\% on A2 (32/90) and from 62.35\% to 68.24\% on A3 (58/85). Adding web retrieval further increases accuracy to 46.67\% (42/90) and 71.76\% (61/85), respectively. The largest gain occurs in A2, with a 20.00 percentage-point improvement over the reported baseline, indicating that sparse numerical reasoning is particularly sensitive to additional guidance and evidence.

However, the benefit of retrieval is not uniform. Compared with Method 1, retrieval corrects 20 previously incorrect A2 predictions but also changes 10 correct predictions to incorrect ones; the corresponding numbers for A3 are 10 and 7. At the schema level, earthquake-depth accuracy increases from 1/10 to 6/10, whereas financial-loss and resilience-score accuracy decreases from 1/5 to 0/5 and from 4/10 to 2/10, respectively. These contrasting outcomes indicate that access to external information alone does not ensure better predictions.
A closer examination of the retrieval records provides further insight. In 32 A2 instances where the retrieval stage reports identifying a specific record containing a target value, correct predictions increase from 12 to 22. In the remaining 58 instances, both methods correctly answer 20 cases in aggregate. 
While the retrieved matches have not been fully verified, the observed improvements demonstrate the potential of external knowledge to support prediction from sparse observations.

\textbf{Implications.}
These findings expose two related limitations in sparse numerical reasoning: difficulty inferring reliable relationships from limited observations and difficulty using external evidence consistently. Explicit priors and retrieval can alleviate these limitations, but their effectiveness depends on the underlying target and the quality of evidence alignment. Importantly, the observed gains reflect improved prediction under additional guidance and knowledge access, not necessarily stronger intrinsic numerical reasoning, since retrieved records may directly contain the target values.

\section{Discussions and Limitations}
\label{DL}

\textbf{Contribution to the community.}
CivilBench advances civil-engineering VLM evaluation from generic image--text understanding to capability-level diagnosis, revealing weaknesses in sparse numerical reasoning, ordinal comparison, and knowledge-grounded decision making. Supplementary experiments show that numerical priors and external knowledge can improve prediction accuracy, but retrieval may also introduce errors when evidence is poorly aligned with the target. These findings distinguish access to information from the ability to use it reliably. For VLM researchers, CivilBench emphasizes domain-specific diagnostics beyond aggregate accuracy; for civil engineers, it provides a basis for assessing model readiness for inspection, risk assessment, and decision support.

\textbf{Limitations.}
CivilBench's \textit{multiple-choice format simplifies real engineering practice}, where explanations, evidence, and uncertainty are essential, while its pseudo-temporal sequences \textit{assess sparse structured reasoning rather than conventional time-series forecasting}. The supplementary experiments are limited to GPT-5-mini, and retrieval may directly provide target values rather than improve intrinsic reasoning. Future work should broaden structural types, damage modes, and regional hazards; incorporate component-level grounding, causal explanations; and distinguish closed-book reasoning from retrieval-augmented prediction.
CivilBench reveals persistent limitations of current VLMs in civil-engineering tasks, particularly in sequential reasoning, sparse numerical inference, and reliable use of domain-specific evidence. 
Future work will expand the benchmark with more diverse engineering scenarios, fine-grained annotations, and open-ended tasks to further assess reasoning reliability and evidence-grounded decision-making.

\section{Conclusion}
\label{Con}

In this work, we introduce CivilBench, a capability-oriented benchmark for evaluating vision-language models in civil and structural engineering. CivilBench organizes engineering tasks into three categories: Sequential Reasoning, Image-based Structural Understanding, and Open-world Understanding, enabling both overall performance evaluation and fine-grained capability diagnosis. Experiments across diverse VLMs reveal substantial performance gaps, with Sequential Reasoning emerging as a major bottleneck, particularly when models must infer relational, ordinal, and temporal patterns from sparse structured observations. Supplementary experiments show that numerical priors and external knowledge can improve prediction accuracy, although reliable use of retrieved evidence remains challenging. These findings indicate that strong visual recognition and general knowledge do not necessarily translate into reliable engineering reasoning. Addressing these limitations requires stronger structured reasoning and effective integration of domain knowledge. CivilBench provides a foundation for evaluating progress toward reliable multimodal decision support in civil engineering.

\section*{Funding}


\printcredits

\bibliographystyle{cas-model2-names}

\bibliography{cas-refs}



\end{document}